\documentclass[letterpaper]{article}
\usepackage{aaai24}
\usepackage{times}
\usepackage{helvet}
\usepackage{courier}
\usepackage[hyphens]{url}
\usepackage{graphicx}
\usepackage{natbib}
\usepackage{caption}
\usepackage{algorithm}
\usepackage{algorithmic}

\usepackage{newfloat}
\usepackage{listings}
\DeclareCaptionStyle{ruled}{labelfont=normalfont,labelsep=colon,strut=off}
\floatstyle{ruled}
\newfloat{listing}{tb}{lst}{}
\floatname{listing}{Listing}
\usepackage{multirow}

\title{Towards a Transformer-Based Reverse Dictionary Model for Quality Estimation of Definitions}
\author {
    Julien Guité-Vinet,
    Alexandre Blondin Massé,
    Fatiha Sadat
}
\affiliations {
     Université du Québec à Montréal, Canada H2X 3Y7\\
    guite-vinet.julien@courrier.uqam.ca, blondin\_masse.alexandre@uqam.ca, sadat.fatiha@uqam.ca
}

\begin{document}

\maketitle

\begin{abstract}
  In the last years, several variants of transformers have emerged.
  In this paper, we compare different transformer-based models for solving the \textit{reverse dictionary task} and explore their use in the context of a serious game called \textit{The Dictionary Game}.
\end{abstract}

In its simplest form, a common language dictionary can be seen as an association between the meaning of a word and its definition.
A task related to word sense disambiguation is called \textit{the reverse dictionary task}.
It focuses on the reverse association, \textit{i.e.} guessing a word from its definition.
A model capable of effectively solving this task should capture multiple semantic relationships such as synonymy, polysemy and homography.
In order to solve the \textit{reverse dictionary task}, some studies rely on information retrieval through lexical database \cite{wordnet-rd,database-driven-rd} or on graph-based approaches \cite{node-rd}.
Recurrent networks have also been shown to be expressive enough to take them into account \cite{cfnn-rd,multi-sense,multi-rd-2020}.
More recent studies have proposed transformer-based models \cite{cross-lingual-rd,unified-model-rd,semeval,t5-rd,data-augmentation}.

In this paper, we investigate different processing of the task with XLNet \cite{xlnet}, the distilled version of BERT \cite{bert} and GPT-2 \cite{gpt}.
We also examine how dictionaries can be semantically developed.
To do so, we experiment with a serious game called \textit{The Dictionary Game} which aims at better understanding the mental lexicon \cite{game}.
It starts with a given root word that the player must define.
The player must recursively define all new words in the definitions, until all words are defined.
Then, we establish a qualitative measure of a dictionary using the fine-tuned models.

\textbf{The Transformer-based Reverse Dictionary.}
The objective of the \textit{reverse dictionary task} is to build a model returning a set of candidates that maximize the probability that a word is paired with a dictionary definition.
The model takes as input a definition in natural language for a word that a user can think of.
It outputs a ranked set of word associated with the given definition.
The Bidirectional Encoder Representations from Transformers (BERT), is a language model that learns from bidirectional representations of unlabeled text by jointly conditioning left and right contexts across all layers \cite{bert}.
The Generative Pre-trained Transformer 2 (GPT-2) is another large language model pretrained on a very large corpus of English data. It is describes as an autoregressive transformer language model consisting of a stack of decoders \cite{gpt}.
Decoders differ from encoders having a masked self-attention mechanism \cite{attention-is-all-you-need}.
XLNet is another autoregressive and bidirectional model \cite{xlnet}.
Instead of using a fixed token order, XLNet considers all possible permutations of the tokens.

\textbf{Methodology.}
We retrieved pretrained models of the DistilBERT, DistilGPT-2 and XLNet transformers.
The distilled versions retain most of the knowledge of the original versions while reducing the number of parameters \cite{distilbert}.
We pooled their outputs, \textit{i.e} a context representation of the input sentence and connected them to a dropout layer followed by a softmax layer.
For DistilBERT, we used the output of the \texttt{[CLS]} special token, which embeds the input sentence for classification tasks.
For DistilGPT-2, we used the output of the last token since that token learns about the previous tokens given the nature of the decoder's attention mechanism.
As for XLNet, we used the average of the hidden states of all tokens in the output layer, excluding the tokens associated with the padding.

All models have been trained using the cross-entropy loss function and the AdamW optimizer.
We added an auxiliary task following the same configuration as the previous task in order to forecast the POS of the target word.
Besides, few definitions from the data exceed 50 tokens after being tokenized.
Therefore, we limited the size of the input sentence, discarding word-definition pairs whose definition exceeds 50 tokens.

\textbf{Evaluations.}
We evaluated our model on the data collected by Hill et al.\ according to three metrics: the median rank of the target words, the rate of occurrences of the target words in the top 1/10/100, and the rank variance of the target word.
The data and metrics are used as a benchmark in several other studies \cite{embedding,cfnn-rd,multi-rd-2020,t5-rd}.

According to our results, XLNet seems to better generalize the task, followed by DistilBERT, then DistilGPT-2.
On the description test set, XLNet forecasts a median rank of 3, a top 1/10/100 of .38/.65/.85 with a rank variance of 364, slightly upgrading the top-1 forecast of previous state of the art models that we are aware of.
Since XLNet and DistilBERT are bidirectional models, this could be a factor improving the accuracy for this task.

Furthermore, one possible application of our models is to evaluate the quality of a dictionary by its definitions.
A definition can be qualified as \emph{perfect}, \emph{good}, \emph{mediocre} if it is in the top-1, top-10, top-100 forecasts respectively, or \emph{wrong} if it lies outside.
One way to get a more meaningful value is to consider the rank or the degree of certainty associated with the model's forecast for a given target word and definition.
We noticed that DistilBERT tends to forecasts better average median ranks.
Since the dictionaries from the data have fairly short definitions, this may have the effect of favoring bidirectional models.

\textbf{Discussion.}\label{sec:discussion}
The degree of certainty seems to be inadequate to assess the quality of a dictionary since for \textit{Merriam-Webster’s}, DistilBERT forecasts an average degree of certainty of 0.55 while averaging the median rank at 0. This is mainly because the model considers synonyms in its probability analysis.

The average rank appears to be a more suitable metric.
However, it can be subject to high variances when a good rank for a definition is assigned to a word of a small dictionary.
It appears that the quality of a dictionary should depend on its size, as well as on the allowed vocabulary, according to the principle that by using several words, we can build deeper lexical semantics.
However, this is more intricate since we have observed that a dictionary having less than 24 words can have a better average rank than a dictionary having more than 110 words.

\textbf{Conclusion.}\label{sec:conclusion}
We have introduced a framework that simulates natural language understanding through a transformer-based model to solve the \textit{reverse dictionary task} and evaluate the quality of dictionaries.
For those tasks, it suggests that performance gains are possible and offers insights into what works when involving transformer-based models and dictionary-based data sets.
Also, dictionary definitions have a very specific grammatical structure, presumably more simpler than the general case of free text.
Therefore, we consider adding more data sources such as urban dictionaries and crossword dictionaries.

\bibliography{aaai24}

\begin{thebibliography}{18}
\providecommand{\natexlab}[1]{#1}

\bibitem[{Blondin~Massé et~al.(2008)Blondin~Massé, Chicoisne, Gargouri,
  Harnad, Picard, and Marcotte}]{game}
Blondin~Massé, A.; Chicoisne, G.; Gargouri, Y.; Harnad, S.; Picard, O.; and
  Marcotte, O. 2008.
\newblock How is meaning grounded in dictionary definitions?
\newblock In \emph{Proceedings of the 3rd Textgraphs Workshop on Graph-Based
  Algorithms for Natural Language Processing, TextGraphs 2008}.

\bibitem[{Chen and Zhao(2022)}]{unified-model-rd}
Chen, P.; and Zhao, Z. 2022.
\newblock A Unified Model for Reverse Dictionary and Definition Modelling.
\newblock In \emph{Proceedings of AACL-IJCNLP 2022}.

\bibitem[{Devlin et~al.(2019)Devlin, Chang, Lee, and Toutanova}]{bert}
Devlin, J.; Chang, M.-W.; Lee, K.; and Toutanova, K. 2019.
\newblock BERT: Pre-training of deep bidirectional transformers for language
  understanding.
\newblock In \emph{Proceedings of the NAACL HLT 2019}, volume~1.

\bibitem[{El-Kahlout and Oflazer(2004)}]{wordnet-rd}
El-Kahlout, I.~D.; and Oflazer, K. 2004.
\newblock Use of wordnet for retrieving words from their meanings.
\newblock In \emph{Proceedings of the GWC 2004}.

\bibitem[{Hill et~al.(2016)Hill, Cho, Korhonen, and Bengio}]{embedding}
Hill, F.; Cho, K.; Korhonen, A.; and Bengio, Y. 2016.
\newblock Learning to understand phrases by embedding the dictionary.
\newblock \emph{Transactions of the Association for Computational Linguistics},
  4.

\bibitem[{Kartsaklis, Pilehvar, and Collier(2018)}]{multi-sense}
Kartsaklis, D.; Pilehvar, M.; and Collier, N. 2018.
\newblock Mapping text to knowledge graph entities using multi-sense LSTMs.
\newblock In \emph{Proceedings of the EMNLP 2018}.

\bibitem[{Li et~al.(2022)Li, Weng, Xia, He, Sun, and Li}]{semeval}
Li, B.; Weng, Y.; Xia, F.; He, S.; Sun, B.; and Li, S. 2022.
\newblock LingJing at SemEval-2022 task 1: Multi-task self-supervised
  pre-training for multilingual reverse dictionary.
\newblock In \emph{Proceedings of SemEval 2022}.

\bibitem[{Mane et~al.(2022)Mane, Patil, Madaswar, and Sadavarte}]{t5-rd}
Mane, S.~B.; Patil, H.~N.; Madaswar, K.~B.; and Sadavarte, P.~N. 2022.
\newblock WordAlchemy: a transformer-based reverse dictionary.
\newblock In \emph{2nd International CONIT}. IEEE.

\bibitem[{Morinaga and Yamaguchi(2018)}]{cfnn-rd}
Morinaga, Y.; and Yamaguchi, K. 2018.
\newblock Improvement of Reverse Dictionary by Tuning Word Vectors and Category
  Inference.
\newblock \emph{Communications in Computer and Information Science}, 920.

\bibitem[{Newman and Aydin(2022)}]{data-augmentation}
Newman, B.; and Aydin, A. 2022.
\newblock Tip of Your Tongue: Methods for an Effective Reverse Dictionary
  Model.

\bibitem[{Radford et~al.(2018)Radford, Narasimhan, Salimans, and
  Sutskever}]{gpt}
Radford, A.; Narasimhan, K.; Salimans, T.; and Sutskever, I. 2018.
\newblock Improving language understanding with unsupervised learning.
\newblock OpenAI.

\bibitem[{Sanh et~al.(2019)Sanh, Debut, Chaumond, and Wolf}]{distilbert}
Sanh, V.; Debut, L.; Chaumond, J.; and Wolf, T. 2019.
\newblock DistilBERT, a distilled version of BERT: smaller, faster, cheaper and
  lighter.
\newblock \emph{arXiv}.

\bibitem[{Shaw et~al.(2013)Shaw, Datta, VanderMeer, and
  Dutta}]{database-driven-rd}
Shaw, R.; Datta, A.; VanderMeer, D.; and Dutta, K. 2013.
\newblock Building a Scalable Database-Driven Reverse Dictionary.
\newblock \emph{IEEE Transactions on Knowledge and Data Engineering}, 25(3).

\bibitem[{Thorat and Choudhari(2016)}]{node-rd}
Thorat, S.; and Choudhari, V. 2016.
\newblock Implementing a Reverse Dictionary, based on word definitions, using a
  Node-Graph Architecture.
\newblock In \emph{Proceedings of COLING 2016}.

\bibitem[{Vaswani et~al.(2017)Vaswani, Shazeer, Parmar, Uszkoreit, Jones,
  Gomez, Kaiser, and Polosukhin}]{attention-is-all-you-need}
Vaswani, A.; Shazeer, N.; Parmar, N.; Uszkoreit, J.; Jones, L.; Gomez, A.~N.;
  Kaiser, {\L}.; and Polosukhin, I. 2017.
\newblock Attention is all you need.
\newblock \emph{Advances in neural information processing systems}, 30.

\bibitem[{Yan et~al.(2020)Yan, Li, Qiu, and Deng}]{cross-lingual-rd}
Yan, H.; Li, X.; Qiu, X.; and Deng, B. 2020.
\newblock BERT for Monolingual and Cross-Lingual Reverse Dictionary.
\newblock In \emph{Proceedings of the EMNLP 2020}.

\bibitem[{Yang et~al.(2019)Yang, Dai, Yang, Carbonell, Salakhutdinov, and
  Le}]{xlnet}
Yang, Z.; Dai, Z.; Yang, Y.; Carbonell, J.; Salakhutdinov, R.~R.; and Le, Q.~V.
  2019.
\newblock Xlnet: Generalized autoregressive pretraining for language
  understanding.
\newblock \emph{Advances in neural information processing systems}, 32.

\bibitem[{Zhang et~al.(2020)Zhang, Qi, Liu, Wang, Liu, and Sun}]{multi-rd-2020}
Zhang, L.; Qi, F.; Liu, Z.; Wang, Y.; Liu, Q.; and Sun, M. 2020.
\newblock Multi-Channel Reverse Dictionary Model.
\newblock \emph{Proceedings of the AAAI Conference on Artificial Intelligence},
  34(01).

\end{thebibliography}
\appendix

\section*{Appendix}
\begin{table*}[h]
  \centering
  \small
  \begin{tabular}{c|c|c}
    Case & Input & Expected Output\\
    \hline
    specific & fruit of the apple tree     &  apple, McIntosh, Cortland \\
    \hline
    antonymous & opposite of opposite        &  same, identical, analogous \\
    \hline
    hyponymous & vehicle carrying passengers &  bus, train, plane \\
    \hline
    hypernymous & lion or tiger or cat        &  animal, panthera, feline\\
    \hline
    ambiguous & a red fruit                 &  apple, cherry, strawberry  \\
    \hline
    polysemous & piece of cake               &  dessert, easy, sugar \\
    \hline
    vague & a living thing              &  animal, human, baby \\
    \hline
    ordered & between 9pm and 6am         &  night, nightfall, midnight\\
    ordered & between 6am and 9pm         &  day, daylight, daytime\\
    \hline
  \end{tabular}
  \caption{Example of inputs and outputs with candidates size $k=3$.
  Given a user input in natural language, the model outputs a set of $k$ candidates.
  The goal is to match the user’s expected word with the model output.
  To effectively disambiguate the definition’s meaning, models have to be able to capture polysemous, antonymous, hyponymous among other semantic relations between the words.}
  \label{tab:rd-input}
\end{table*}

\begin{figure}[h]
  \centering
  \includegraphics[width=0.3\textwidth]{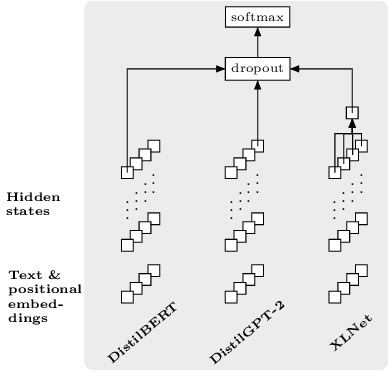}
  \caption{Pooled outputs for the DistilBERT, DistilGPT-2 and XLNet architectures.
  Considering the last hidden state of a transformer for an input sentence, we use the \texttt{[CLS]} token for DistilBERT, the last token for DistilGPT-2 and an average pooling for XLNet.}
  \label{fig:model}
\end{figure}

\begin{table}[h]
  \centering
  \scriptsize
  \begin{tabular}{|c | c|}
    \hline
    training framework & tensorflow \\
    batch size & 128\\
    learning rate & 1e-4\\
    optimizer & AdamW \\
    beta1, beta2 & 0.9, 0.999\\
    weight decay & 1e-6\\
    word loss & cross-entropy\\
    tag loss & cross-entropy\\
    Transformer depth & 12\\
    Transformer heads & 12\\
    Transformer dropout & 0.1\\
    Transformer output dropout & 0.4\\
    maximum sequence length & 50 tokens\\
    \hline
  \end{tabular}
  \caption{
    Our model configurations is summarized here.
    On a single Nvidia Tesla V100, it takes 160 hours for a model to complete 200 epochs on the reverse dictionary task.
  }
  \label{tab:hyperparameters}
\end{table}

\begin{figure}[t]
  \centering
  \tiny
  \begin{tabular}{|ll|ll|}
                          \hline
                          Word & Definition & Word & Definition \\
                          \hline
                          horse & animal leg        & animal & alive thing \\
                          alive & not dead          & leg & useful thing \\
                          useful & good             & dead & not alive  \\
                          thing  &  thing           & not   & not \\
                          good & not bad            & bad   & not good \\
                          dog & animal leg          &    &  \\
                          \hline
  \end{tabular}
  \includegraphics[width=0.23\textwidth]{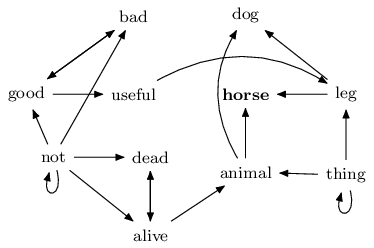}
  \caption{
    A dictionary represented as a table (top) and as a directed graph (bottom).
    The root word \textit{horse} is identified in bold.
    Stop words have been removed from the definitions and remaining words have been lemmatized.
    When the game ends, the directed graph has the property that for each vertex there is a path to the root word.
  }
  \label{fig:game}
\end{figure}

\begin{figure}[h]
  \centering
  \includegraphics[width=0.24\textwidth]{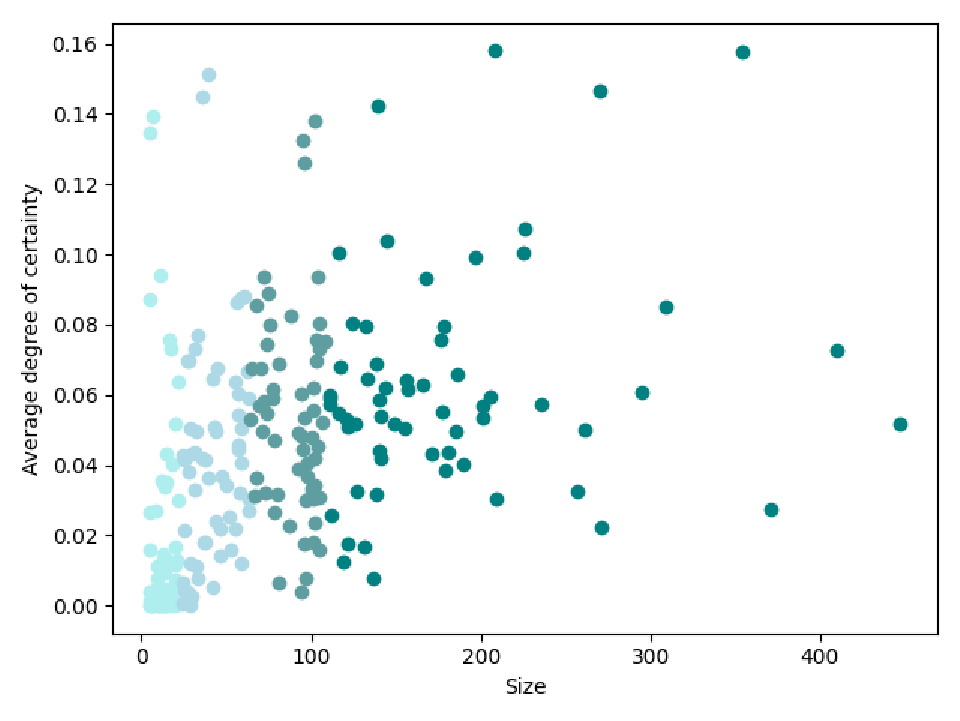}
  \includegraphics[width=0.24\textwidth]{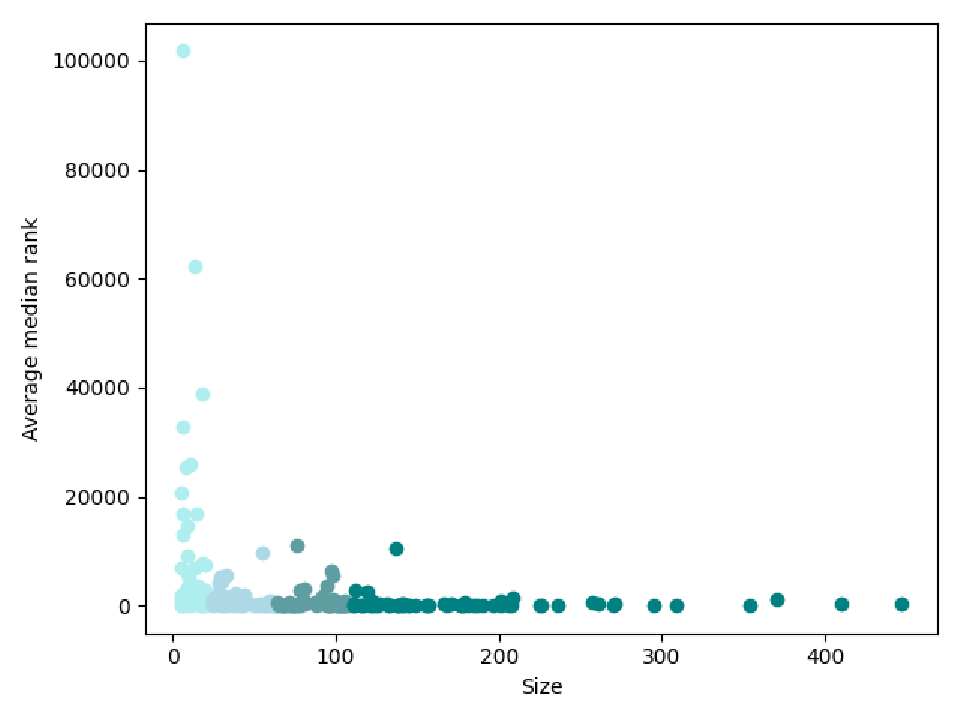}
  \caption{Each color represents a different quartile where a dictionary stands.
  The larger the vocabulary of a dictionary is, the better the models tend to forecast with respect to the average degree of certainty (top) and the average median rank (bottom).}
  \label{fig:scatter}
\end{figure}

\begin{table*}[h]
  \centering
  \small
  \begin{tabular}{|c|ccc|ccc|ccc|}
    \hline
    Model         &         & Seen Definition   &             &   &  Unseen Definition  &   &  &  Description    &     \\
    \hline
    OneLook            &  \textbf{0} & .66/.94/.95          & 200         & - & - & -                      & 5.5    & .33/.54/.76  & 332      \\
    \hline
    BoW           & 172        & .03/.16/.43          & 414         & 248 & .03/.13/.39 & 424        & 22     & .13/.41/.69  & 308  \\
    RNN           & 134        & .03/.16/.44          & 375         & 171 & .03/.15/.42 & 404        & 17     & .14/.40/.73  & 274  \\
    BiLSTM        & 25         & .18/.39/.63          & 363         & 101 & .07/.24/.49 & 401        & 5      & .25/.60/.83  & 214  \\
    MS-LSTM       & \textbf{0} & \textbf{.92}/.98/.99 & 65          & 276 & .03/.14/.37 & 426        & 1000   & .01/.04/.18  & 404  \\
    Multi-channel &  16        & .20/.44/.71          & 200         & 54  & .09/.29/.58 &  \textbf{358} & 2   & .32/.64/.88  & 203  \\
    BERT(MLM)     &  \textbf{0}& .57/.86/.92          & 240         & 18  & .20/.46/.64     & 418    & \textbf{1}   & 36/.66/.94 & \textbf{94} \\
    Unified       &  -         & -                    & -           & 18  & .13/.39/\textbf{.81}     & 386    & 4    & .22/.64/\textbf{.97} & 183 \\
    T5            &  -         & .55/.88/.95          & -           & -   & .14/\textbf{.58}/.78              & -      & -    & .28/\textbf{.88}/.96 & - \\
    \hline
    DistilBERT(CLS) & \textbf{0} & .77/\textbf{.99/1.0}& \textbf{3}     & \textbf{15} & \textbf{.24}/.45/.64 & 475    & 4      & .31/.60/.80 & 397 \\
    DistilGPT-2     & \textbf{0} & .79/\textbf{.99/1.0} & \textbf{3}    & 41          & .18/.40/.57          & 488    & 4      & .31/.57/.78 & 414  \\
    XLNet           & \textbf{0} & .77/\textbf{.99/1.0} & \textbf{3}    & \textbf{15} & \textbf{.23/.46}/.64 & 469    & 3      & \textbf{.38}/.65/.85 & 364 \\
    \hline
    \multicolumn{4}{c}{}  & \multicolumn{6}{|c|}{median rank \quad accuracy@1/10/100 \quad rank variance}  \\
  \end{tabular}
  \caption{
    The results of our experimentation on the DistilBERT(CLS), DistilGPT-2 and XLNet transformers adapted to the \textit{reverse dictionary task}.
    Our results are compared with the results taken from the papers of Yan et al. \cite{cross-lingual-rd} and Chen et Zhao. \cite{unified-model-rd}.
    For each set of test data, three columns represent respectively the median rank, the accuracy@1/10/100 and the rank variance.
    Chen and Zhao omit the seen definition results since they state that the goal of the model is to generalize unseen data.
  }\label{tab:results}
\end{table*}

\begin{table*}[h]
  \centering
  \small
  \begin{tabular}{|c|cccc|c|}
    \hline
    Models & Q1 & Q2 & Q3 & Q4 & Merriam-Webster’s  \\
    \hline
    DistilBERT &  .020/20013/1576 & .040/12712/382 & .053/11275/208 & .062/\textbf{9710}/\textbf{127}   & .55/12/0 \\
    DistilGPT-2 & .021/17268/2199  & .035/12177/722 & .047/10591/498 & .056/9828/311 &  .58/12/0 \\
    XLNet & .022/26377/2971  & .044/17427/669 & .056/15070/286 & \textbf{.068}/12482/133   & .53/12/0\\
    \hline
    \multicolumn{3}{c}{}  & \multicolumn{3}{|c|}{certainty \quad avg rank \quad median rank}  \\
  \end{tabular}
  \caption{
    The quality of a dictionary is evaluated according to the average degree of certainty/rank/median rank.
    We aggregate the results conforming to the quartile where a dictionary is found, \textit{i.e.} \textit{Q1} regroups the results of dictionaries with less than 24 words, \textit{Q2} between 24 and 63, \textit{Q3} between 64 and 110 and \textit{Q4} greater than 110 words.
  }
  \label{tab:evaluation-games}
\end{table*}

\begin{table*}[h]
  \centering
  \scriptsize
  \begin{tabular}{|c|cc|ccc|}
    \hline
    Definition & target word & target POS & model & forecast word  & forecast POS \\
    \hline
    & & & DiGPT-2 & chocolate  & NN\\
    & &  & XLNet & mocha  & NN\\
    \multirow{-3}{20em}{chocolate that contains at least 32 percent cocoa butter                                                                          }& \multirow{-3}{2em}{couverture} & \multirow{-3}{2em}{NN}  & DiBERT & cacao & NN\\
    \hline
    & & & DiGPT-2 & paint  & NN\\
    & &  & XLNet & stuffing  & NN\\
    \multirow{-3}{20em}{a material used to coat cooking utensils and in industrial applications where sticking is to be avoided                           }& \multirow{-3}{3em}{polytetra fluoro ethylene} & \multirow{-3}{2em}{NN} & DiBERT & dressing & NN \\
    \hline
    & & & DiGPT-2 & lifestyle  & NN\\
    & &  & XLNet & politics  & NN\\
    \multirow{-3}{20em}{military actions designed to influence the perceptions and attitudes of individuals, groups, and foreign governments              }& \multirow{-3}{2em}{psyop} & \multirow{-3}{2em}{NN} & DiBERT & culture & NN \\
    \hline
    & & & DiGPT-2 & cytogenetics  & NN\\
    & &  & XLNet & genetics  & NN\\
    \multirow{-3}{20em}{the branch of genetics that studies the genetically determined variations in responses to drugs in humans or laboratory organisms }& \multirow{-3}{2em}{pharmaco genetics} & \multirow{-3}{2em}{NN}  & DiBERT & genetics & NN\\
    \hline
    & & & DiGPT-2 & ranch  & NN\\
    & &  & XLNet & ranch  & NN\\
    \multirow{-3}{20em}{a small town in a cattle, raising area of western North America                                                                   }& \multirow{-3}{2em}{cowtown} & \multirow{-3}{2em}{NN} & DiBERT & pueblo & NN  \\
    \hline
    & & & DiGPT-2 & randomness  & NN\\
    & &  & XLNet & randomness  & NN\\
    \multirow{-3}{20em}{the quality of lacking any predictable order or plan                                                                              }& \multirow{-3}{2em}{stochasticity} & \multirow{-3}{2em}{NN}  & DiBERT & noise & NN \\
    \hline
  \end{tabular}
  \caption{
    A short list of some errors made by our models when the target word does not appears in the top 5000 candidates.
    One of the models must also have made a separate forecast from the other two.
    The target POS of the table include singular common nouns (\texttt{NN}), past participle verbs (\texttt{VBN}) and adjectives (\texttt{JJ}).
    One limitation of our model is that they push some very technicals terms, that are initaly related to other more commun words, out of their shared vector space.
    We argue that specialized terminalogy are less frequent in usual dictionaries, and adding more specialized terms in the dataset could help alievate the problem.
  }
  \label{tab:error-analisis}
\end{table*}

\end{document}